%% file: main.tex
\documentclass[10pt,twocolumn,letterpaper]{article}

\usepackage{iccv}              % To produce the CAMERA-READY version
\usepackage[accsupp]{axessibility}  % Improves PDF readability for those with disabilities.
\renewcommand{\paragraph}[1]{\vspace{1.25mm}\noindent\textbf{#1}}

\input{preamble}

\definecolor{iccvblue}{rgb}{0.21,0.49,0.74}
\usepackage[pagebackref,breaklinks,colorlinks,allcolors=iccvblue]{hyperref}

\def\paperID{12850} % *** Enter the Paper ID here
\def\confName{ICCV}
\def\confYear{2025}

\title{A Quality-Guided Mixture of Score-Fusion Experts Framework \\ for Human Recognition}

\author{
Jie Zhu, Yiyang Su, Minchul Kim, Anil Jain, and Xiaoming Liu \\
Department of Computer Science and Engineering, \\
Michigan State University, East Lansing, MI 48824 \\
{\tt\small \{zhujie4, suyiyan1, kimminc2, jain, liuxm\}@msu.edu}
}
\begin{document}
\maketitle

%%%%%%%%% ABSTRACT
\begin{abstract}
  Whole-body biometric recognition is a challenging multimodal task that integrates various biometric modalities, including face, gait, and body. This integration is essential for overcoming the limitations of unimodal systems. Traditionally, whole-body recognition involves deploying different models to process multiple modalities, achieving the final outcome by score-fusion (e.g., weighted averaging of similarity matrices from each model). However, these conventional methods may overlook the variations in score distributions of individual modalities, making it challenging to improve final performance. In this work, we present \textbf{Q}uality-guided \textbf{M}ixture of score-fusion \textbf{E}xperts (QME), a novel framework designed for improving whole-body biometric recognition performance through a learnable score-fusion strategy using a Mixture of Experts (MoE). We introduce a novel pseudo-quality loss for quality estimation with a modality-specific Quality Estimator (QE), and a score triplet loss to improve the metric performance. Extensive experiments on multiple whole-body biometric datasets demonstrate the effectiveness of our proposed approach, achieving state-of-the-art results across various metrics compared to baseline methods. Our method is effective for multimodal and multi-model, addressing key challenges such as model misalignment in the similarity score domain and variability in data quality. Code is available at the \href{https://github.com/jiezhu23/QME_ICCV25}{Project Link}.
\end{abstract}

%%%%%%%%% BODY TEXT
\input{1-introduction}

\input{2-relatedworks}

\input{3-method}

\input{4-experiments}

\section{Conclusion}

We propose \textbf{QME}, a framework for whole-body biometric recognition that dynamically fuses modality-specific experts through a novel quality-aware weighting. To enhance discriminative power, we introduce a score triplet loss that explicitly enforces a margin between match and non-match scores. Experiments across diverse benchmarks demonstrate the superior performance of our method, serving as a general framework for multi-modal score fusion, which can be applied to any system with heterogeneous models. 

% We propose \textbf{QME}, a unified framework for whole-body biometric recognition that dynamically fuses modality-specific experts through a novel quality-aware weighting mechanism. To enhance discriminative power, we introduce a score triplet loss that explicitly enforces a margin between match and non-match scores. Extensive experiments across diverse benchmarks demonstrate the superior and consistent performance of our approach. QME serves as a generalizable and plug-and-play solution for multi-modal score fusion, applicable to a wide range of biometric systems with heterogeneous backbones.

\paragraph{Acknowledgments.}
This research is based upon work supported in part by the Office of the Director of National Intelligence (ODNI), Intelligence Advanced Research Projects Activity (IARPA), via 2022-21102100004. The views and conclusions contained herein are those of the authors and should not be interpreted as necessarily representing the official policies, either expressed or implied, of ODNI, IARPA, or the U.S. Government. The U.S. Government is authorized to reproduce and distribute reprints for governmental purposes notwithstanding any copyright annotation therein.

%%%%%%%%% REFERENCES
{\small
\bibliographystyle{ieee_fullname}
\bibliography{egbib}
}

% \clearpage
% \input{supp}

\end{document}

%% file: preamble.tex
\usepackage{multirow}
\usepackage{pifont}
\usepackage{color, colortbl}

%% file: 1-introduction.tex
\section{Introduction} \label{sec:intro}

\begin{figure}[t!]
    \centering
    \includegraphics[width=1\linewidth]{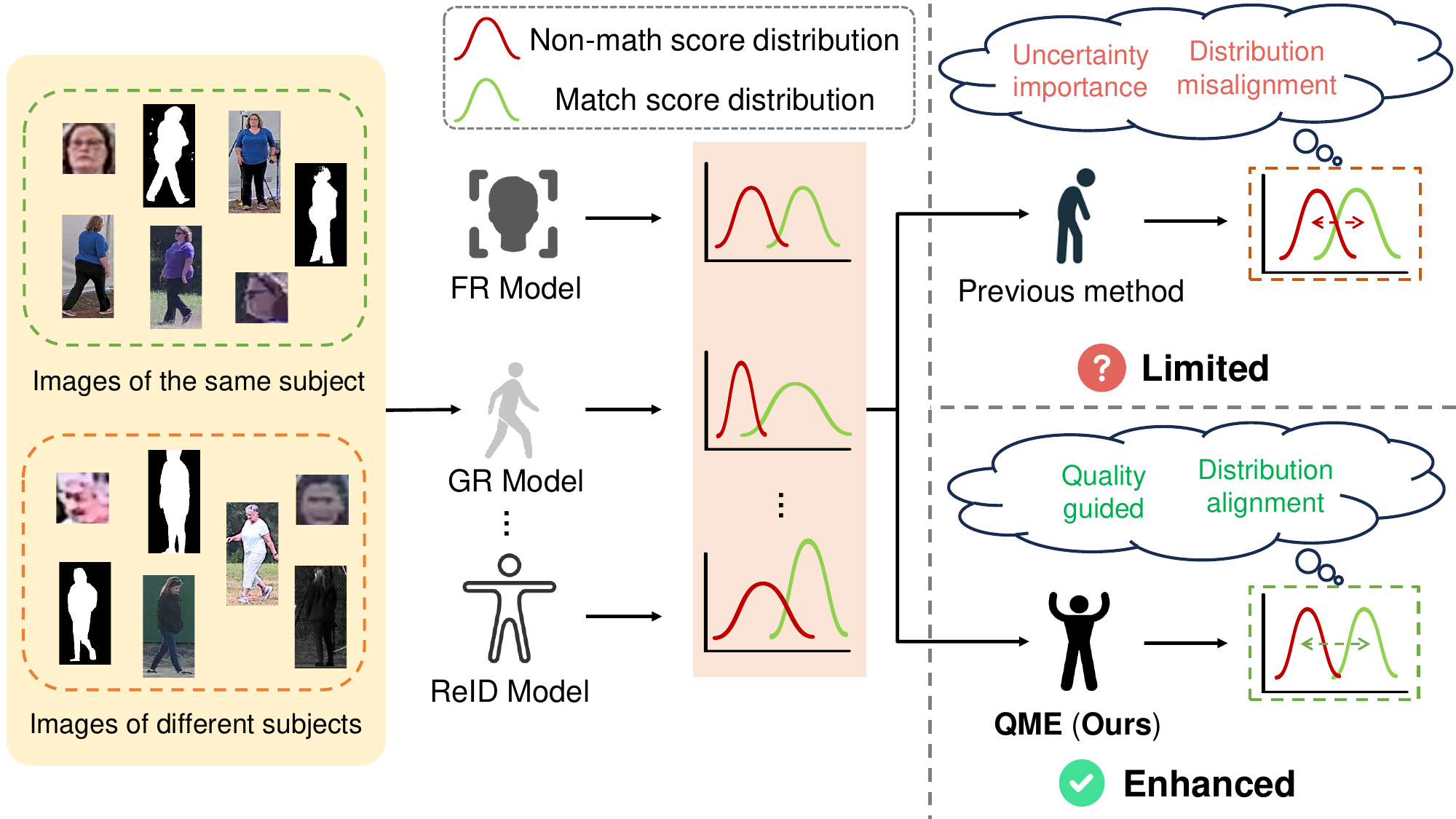}
    \caption{Illustration of score distribution alignment in multimodal human recognition. Different models and modalities (\eg, face, gait, and body) produce distinct similarity score distributions. Conventional score-fusion methods struggle with optimal alignment and assigning importance weights to each modality, potentially degrading performance.}
    \label{fig:motivation}
\end{figure}

Whole-body biometrics integrates diverse recognition tasks such as Face Recognition (FR)~\cite{deng2019arcface, kim2022adaface},  Gait Recognition (GR)~\cite{zhang2019gait, ye2024biggait},  and Person Re-identification (ReID)~\cite{gu2022clothes, liu2024distilling} to overcome unimodal limitations. Whole-body biometrics benefits from the combined strengths of multiple modalities. This multimodal synergy ensures robust performance in non-ideal conditions (low-light, occlusion, and missing traits), making it indispensable for security-critical domains like surveillance and law enforcement. 

Effective fusion is pivotal to whole-body recognition. Current approaches include decision-level fusion, feature-level fusion, and score-level fusion~\cite{singh2019comprehensive}. In decision-level fusion, each modality first makes an identity decision based on its extracted features. The individual decisions are then combined based on either decision scores or ranks. 
% This fusion scheme does not incorporate any correlation among the modalities. 
Feature-level fusion combines extracted features from different modalities to obtain a single representation~\cite{kim2025sapiensid,chen2023atm}. However, this approach is often hindered by inconsistencies across modalities in biometrics, as different traits may not necessarily complement each other effectively. Most importantly, feature-level fusion requires suitable paired multimodal datasets. Many available datasets such as WebFace42M~\cite{zhu2021webface260m} for face recognition do not contain whole-body data, while other datasets like PRCC~\cite{yang2019person}, LTCC~\cite{qian2020long}, and CCPG~\cite{li2023depth} widely used in person ReID and gait recognition, are limited by dataset size, the masking of faces, or insufficient number of subjects for generalizable training.

Compared to feature-level fusion, score-level fusion integrates the similarity scores or feature (embedding) distances generated by individual models. Score-level fusion offers computational efficiency and modular flexibility compared to feature-level fusion, enabling seamless integration of heterogeneous modalities while preserving individual models' performance. However, conventional score-fusion techniques are limited by their inability to fully utilize the different distributions of match (genuine) and non-match (impostor) scores produced by each model, as shown in Fig.~\ref{fig:motivation}. Additionally, finding the optimal weight for each model in the fusion process is challenging, even using grid search~\cite{liashchynskyi2019grid}, leading to suboptimal performance.

To address these challenges, we propose a Quality Estimator (QE) and pseudo-quality loss that leverages pretrained models to generate pseudo-quality labels, eliminating laborious manual annotation. We develop a Mixture of Score-Fusion Experts method, where each expert learns a distinct fusion strategy (\eg, one prioritizes face-gait synergy, and another handles occlusion scenarios). Experts’ contributions are dynamically weighted by QE predictions, ensuring robustness to sensor noise and missing modalities. To improve metric learning performance, we present a score triplet loss that enforces margin separation between match/non-match scores while suppressing non-match magnitudes, directly aligning with metrics like 1:1 verification and 1:N open-set search. This approach improves score-level alignment between modalities without the need for retraining biometric backbones nor requiring tremendous training data. Our main contributions are:

\begin{itemize}[noitemsep, topsep=2pt, leftmargin=*]
    \item We propose a Quality Estimator (QE) that employs pseudo quality loss—derived from pretrained models and ranking performance—to assess biometric modality quality without the need for human-labeled data.
    \item We introduce \textbf{QME}, a multimodal biometric recognition framework that integrates a learnable, modality-specific score-fusion method. QME dynamically combines diverse fusion strategies, adapting to sensor noise, occlusions, and missing modalities.
    \item We introduce a novel score triplet loss for metric learning by enforcing a match/non-match score margin, directly improving key metrics like verification accuracy and open-set search effectiveness.
    \item Experiments on multiple whole-body biometric datasets validate our approach's superior robustness over leading score-fusion methods and models.
\end{itemize}

%% file: 2-relatedworks.tex
\begin{figure*}[t!]
    \centering
    \includegraphics[width=.9\linewidth]{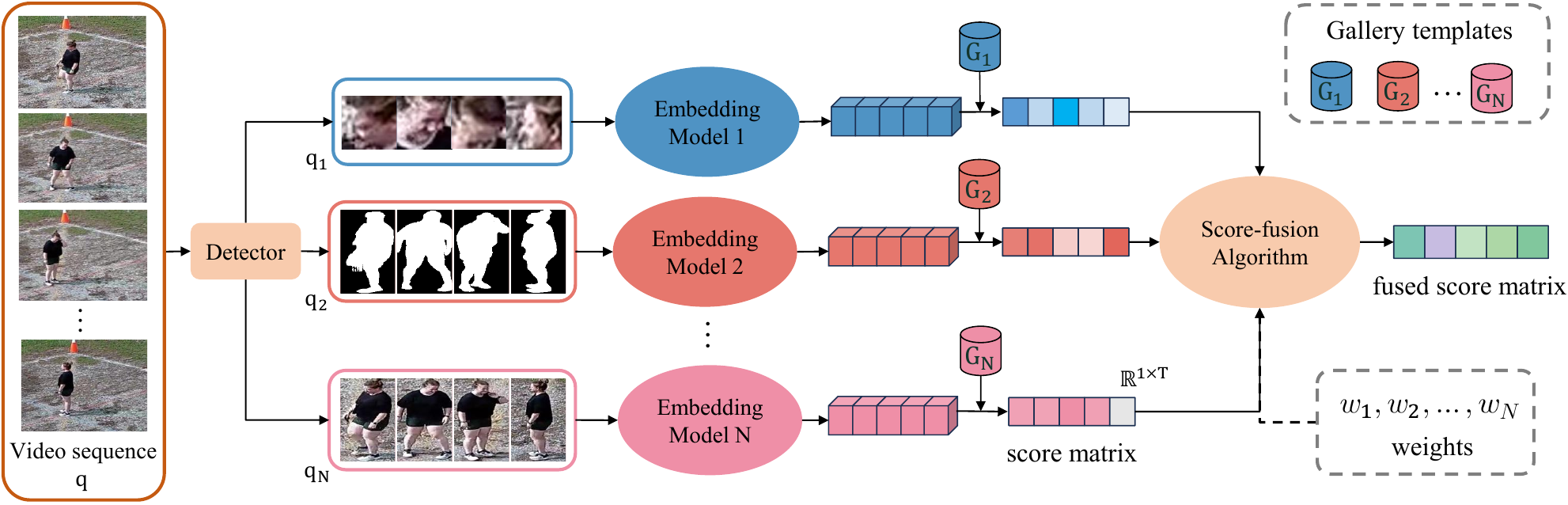}
    \caption{General framework for whole-body biometric recognition. An input video sequence $q$ is processed by a detector to extract different modality queries, which are fed into multiple embedding models. Each model generates similarity scores by comparing the extracted features with $T$ gallery templates. Our work focuses on score-fusion algorithms that produce the final decision based on input score matrices and modality weights.}
    \label{fig:whole_body_recognition}
\end{figure*}

\section{Related Work} \label{sec:related work}

\subsection{Score-fusion}

 Score-level fusion integrates similarity scores from multiple modalities to optimize recognition decisions~\cite{singh2019comprehensive}. Traditional score-fusion methods include Z-score and min-max normalization.~\cite{poh2007improving, nandakumar2007likelihood, vatsa2007integrating, poh2011unified, he2010performance} introduce likelihood ratio based score fusion. Ross \textit{et al.} propose mean, max, or min score-fusion, where the final score is determined by averaging, the highest, or the lowest score~\cite{ross2003information,jain2005score,yilmaz2016score}. 
 Recent literature categorizes score fusion into two paradigms: fixed-rule methods, employing predefined heuristics (\eg, predefined weights), and trained-rule methods, utilizing learned parameters optimized through training (\eg, SVM)~\cite{teng2022optimized, cheniti2024approach, park2021multi}. Score-fusion methods offer several advantages: 1) they are robust to missing modality inputs, and 2) they simplify alignment, as the domain gap between modalities is smaller than feature-space alignment. However, challenges remain in determining the optimal alignment and weighting for each model and identifying the most effective fusion strategy. We aim to explore a better way of assessing the contribution of each modality and develop a more generalizable score-fusion method. 

\subsection{Biometric Quality Assessment}
Unlike generic image quality assessment~\cite{saha2023re}, biometric quality assessment is the process of evaluating the quality of biometric data (\eg, facial images), which directly influences recognition performance~\cite{grother2007performance,ozay2009improving,tong2010improving}. This assessment typically follows initial authentication to filter out spoofed or synthetic samples~\cite{zhang2024common, guo2022multi, guo2025m2f2det}.
While some studies target fingerprints and irises~\cite{el2015quality, bharadwaj2014biometric, krichen2007new}, others apply learning-based methods for facial image quality~\cite{chang2020data, best2018learning, hernandez2019faceqnet, terhorst2020ser, shi2019probabilistic, meng2021magface, kim2022adaface, kim2022cluster}. However, many such methods rely on specialized training procedures incompatible with pretrained models. In this work, we introduce a method to train a general QE by distilling knowledge from the pretrained model, providing a versatile approach to biometric quality assessment.

\subsection{Whole-Body Biometric Recognition}
As illustrated in Fig.~\ref{fig:whole_body_recognition}, whole-body biometric systems integrate detectors, encoders, and fusion modules to unify multi-modal traits (\eg, face, gait) for robust identification~\cite{de2012cabala}. Key to the design is effectively leveraging complementary strengths while mitigating individual weaknesses: facial recognition excels with high-resolution frontal images but degrades under non-ideal conditions (\eg, large standoff, off-angle views), while gait and ReID models contend with clothing/posture variations~\cite{liu2024farsight, liu2025person}. Recent advances~\cite{cornett2023expanding,huang2023whole,guo2023multi,yang2024shallow,ren2024implicit,su2025open} highlight multi-attribute fusion but largely overlook the heterogeneity inherent in whole-body modalities, focusing mainly on homogeneous sensor data. Efforts to incorporate facial features into ReID~\cite{li2016toward,li2017learning,grudzien2018face,liu2024farsight,kim2025sapiensid} often prioritize modular additions over optimizing fusion efficacy. 
Fusion methods for comprehensive whole-body biometric recognition remain challenging, and require in-depth exploration.

%% file: 3-method.tex
\section{Methodology} \label{sec:method}

In this section, we introduce the proposed \textbf{QME} method, which leverages quality assessment and learnable score-fusion with MoE across multiple modalities. Our approach is specifically designed to tackle challenges related to model misalignment in score-level distributions and varying data quality in whole-body biometric recognition.

\paragraph{Overview.} In biometric evaluation, a query (or probe) refers to a sample sequence needing identification/verification against a gallery of enrolled subjects in the system. Each gallery subject may have multiple videos/images to extract gallery templates. Given a model $M_n$ in the embedding model set $\{M_1, M_2, \dots, M_N\}$ with a query and gallery templates where $N$ is the number of models, we compute the query feature $q_n \in \mathbb{R}^{L \times d_n}$ and gallery template features $G_n \in \mathbb{R}^{T \times d_n}$, where $L$ represents the sequence length of the query (number of images) and $T$ is the total number of gallery templates (videos/images) across all gallery subjects, and $d_n$ is the feature dimension of $M_n$. We further compute the average of $q_n$ to obtain the query-level feature vector in $\mathbb{R}^{d_n}$, and then compute its similarity with $G_n$ to get the query score matrix $\mathbf{S}_n \in \mathbb{R}^{1 \times T}$, representing the similarity score of the query with each gallery template. Our training process involves two stages: (1) training QE, and (2) freezing QE while training the learnable score-fusion model. 

\begin{figure*}[t]
    \centering
    \includegraphics[width=.95\linewidth]{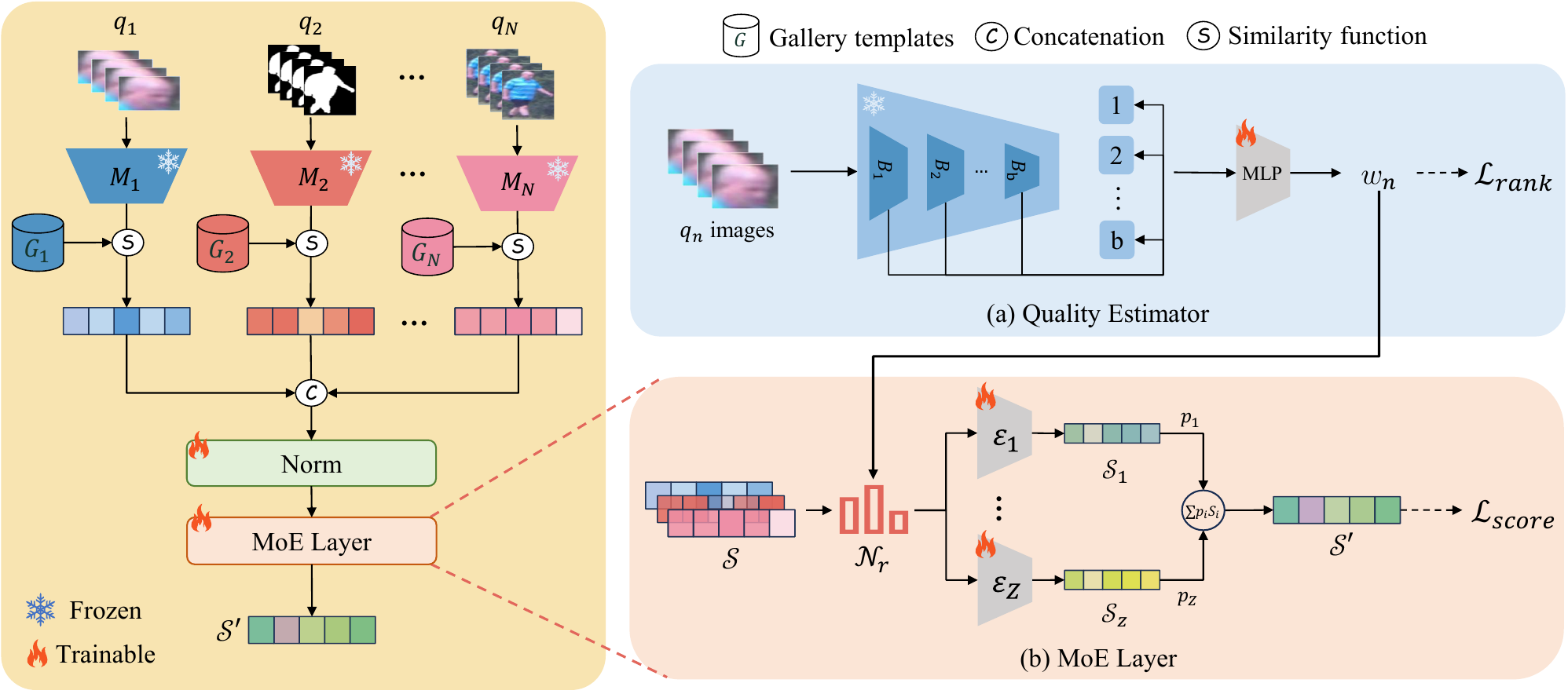}
    \caption{The architecture of the proposed QME framework. It includes a $\mathit{Norm}$ layer and an $\mathit{MoE}$ layer to process concatenated score matrix $\mathcal{S}$ from the model set ${M_1, M_2, \dots, M_N}$. The $\mathit{MoE}$ layer contains experts ${\varepsilon_1, \varepsilon_2, \dots, \varepsilon_Z}$ to individually encode the fused score matrices. A quality estimator (QE) uses the intermediate feature $\mathcal{I}_n$ from the backbone block ${B_1, B_2, \dots, B_b}$ to generate weights $w_n$, which control ${p_1, p_2, \dots, p_Z}$ for a weighted sum, producing the final fused score matrix $\mathcal{S}'$.}
    \label{fig:LSF_MoE}
\end{figure*}

\subsection{Quality Estimator (QE)} \label{subsec: qe}

The goal of the QE is to predict the input quality of a given modality. We hypothesize that if the input quality for a particular modality is poor, the system should shift focus to other modalities to enhance overall performance. As illustrated in Fig.~\ref{fig:LSF_MoE}(a), to train a QE for $M_n$, we collect the intermediate features $\mathcal{I}_n \in \mathbb{R}^{L \times U \times P_n \times d_n}$ from $M_n$, where $U$ is the number of blocks, $P_n$ is the patch size of $M_n$. $\mathcal{I}_n$ captures various levels of semantic information from the model. We follow~\cite{kim2022cluster} to extract intermediate features from the backbone and compute the mean and the standard deviation, reducing $\mathcal{I}_n$ to a representation in $\mathbb{R}^{L \times 2d_n}$. This representation is then fed into an encoder to predict query-level quality weight $w_n \in \mathbb{R}$ produced by sigmoid function.

\paragraph{Pseudo Quality Loss.} The challenge of training QE is the lack of human-labeled qualities. Empirically, we do not have the quality label of the query images. However, we can know the ranking result by sorting the similarities between the query feature and training gallery features. A higher ranking result indicates the input images are close to their gallery center. We assume that if the ranking result of the input is better, the quality of the input will be higher. Hence, we propose a pseudo quality loss $\mathcal{L}_{\mathit{rank}}$ using the ranking result of the input for the pretrained model $M_n$:
\begin{align}
\mathcal{L}_{\mathit{rank}}=\sum_{i \in L} \operatorname{MSELoss}{\left(w_i,\operatorname{ReLU}(\frac{\delta-r_i}{\delta-1})\right)}.
\label{eq:ranking_loss}
\end{align}

Here $r_i$ is the ranking result of the query feature $q_i$, $w_i$ is the predicted quality weight, and $\delta$ is a hyperparameter to adjust the sensitivity of the ranking threshold. To obtain $r_i$, we compute the similarity matrix between $q_i$ and $G_n$. Lower $\delta$ will push the predicted $r_i$ to 0 if the ranking result is out of $\delta$. Conversely, higher $\delta$ will cause the QE to predict a value closer to 1 as it has a higher tolerance for the ranking result.
Our proposed QE offers several benefits: (1) It can generalize across all pretrained models (not only FR models) by learning from these models and identifying characteristics of challenging samples, and (2) it can be trained on any dataset, whether in-domain or out-of-domain. While pretrained models may exhibit biases toward their training data, which can hinder generalization, challenging samples may originate from either in-domain or out-of-domain data. 

\subsection{Mixture of Score-fusion Experts} \label{subsec:moe}
The concept of MoE~\cite{shazeer2017outrageously, fedus2022switch} comes from the NLP community, where they use MoE layers to replace feed-forward network (FFN) layers in the transformer blocks. With the sparsity of experts and the router network, each expert can focus on handling different tokens. In addition, some special loss functions are designed to control the behavior of the router~\cite{shazeer2017outrageously,shazeer2018mesh,lepikhin2020gshard,zuo2021taming,dai2021generalizable,su2025hierarchical}. 

Inspired by this, we design a MoE layer (shown in Fig.~\ref{fig:LSF_MoE}(b)) with multiple score-fusion experts, controlled by $\mathit{\mathcal{N}_r}$ that learns to perform score-fusion based on quality weights. 
Unlike in traditional MoE setups, where a router network predicts assignment probabilities from inputs, the similarity score in our case is a high-level semantic feature, lacks fine-grained cues about query quality. Instead, we use the proposed QE to predict the quality weight of the query to imply the reliability of the input modality, guiding the selection process. For an expert $\epsilon_z$ from expert set $\{\epsilon_1, ..., \epsilon_Z\}$ where $Z$ is the number of experts, it receives a concatenated score matrix $\mathcal{S} \in \mathbb{R}^{T \times N}$ from all modalities and predict a fused score matrix $\mathcal{S}_z \in \mathbb{R}^{1 \times T}$. Given $w_n$ as the modality-specific quality weight and $\varepsilon_n$ controlled by $p_n = w_n$, we aim for expert $\varepsilon_n$ to prioritize the selected modality when $w_n$ is high. Conversely, when $w_n$ is low, other experts contribute more to the final score matrix and shift the focus to other modalities. This approach ensures that higher-quality modalities have a greater influence on the output, while lower-quality ones contribute less, optimizing overall performance. 

\subsection{Quality-Guided Mixture of Score-fusion Experts (QME)} \label{subsec: lsf}

Based on Sec.~\ref{subsec: qe} and~\ref{subsec:moe}, we further introduce QME. As illustrated in Fig.~\ref{fig:LSF_MoE} (left), for a query feature set $\mathbf{Q}=\{q_1, q_2, \dots, q_N\}$ processed by the model set $\{M_1, M_2, \dots, M_N\}$, we generate the concatenated input score matrix $\mathcal{S}=\{\mathbf{S}_1, \mathbf{S}_2, \dots, \mathbf{S}_N\} \in \mathbb{R}^{T \times N}$. For models that use Euclidean distance as a metric, we convert distances into similarity scores:
\begin{align}
\frac{1}{1+\mathit{Euc}(q, g)},
\label{eq:euc_scores}
\end{align} 
where $\mathit{Euc}(q, g)$ represents Euclidean distance between the query feature $q$ and the gallery feature $g$. This transformation remaps Euclidean distances to align with the range of Cosine Similarity, where larger values indicate higher similarity. We then normalize $\mathcal{S}$ using a $\mathit{BatchNorm}$ layer. After normalization, $\mathcal{S}$ is fed into the MoE layer, which contains a router network $\mathit{\mathcal{N}_r}$ and multiple score-fusion experts $\{\varepsilon_1, \varepsilon_2, \dots, \varepsilon_Z\}$. Each expert is specialized to handle specific input conditions (\ie, similarity values), with the router selecting the most suitable expert based on quality assessment. $\mathit{\mathcal{N}_r}$ takes $w_n$ as the input and generates the weight of assigning input to all experts $\{p_1, p_2, \dots, p_Z\}$ where $p_Z$ is the weight of contribution of expert $\varepsilon_Z$.
The final fused score matrix $\mathcal{S}'$ is computed as a weighted sum of the outputs from all experts:
\begin{align}
    \mathcal{S}'=\sum_{z \in Z} p_z \mathcal{S}_z, 
\end{align}
where $\mathcal{S}_z$ is the output score matrix from $\varepsilon_z$. By using quality weight to modulate $\mathcal{S}'$, each expert learns how the contributions of different modalities’ scores to $\mathcal{S}'$ should be adjusted in response to changes in their quality levels.

\paragraph{Score Triplet Loss.}
The triplet loss~\cite{schroff2015facenet} optimizes relative distances between samples:
\begin{align}
\mathcal{L}_{\mathit{tri}} = \operatorname{ReLU}(d(a, p) - d(a, n) + m),
\label{eq:triplet_loss}
\end{align}
where $d(a, p)$ is the distance between anchor $a$ and positive sample $p$, $d(a, n)$ is the distance between anchor $a$ and negative sample $n$, and $m$ enforces a margin. The triplet loss focuses on maintaining a boundary between positive and negative pairs, but it does not effectively constrain the value of non-match scores. The verification and open-set search rely on a threshold $\tau$. For example, TAR@$\tau\%$FAR measures the acceptance rate of the match samples such that only $\tau\%$ of non-match scores can be accepted as matches. To optimize these metrics, we introduce the score triplet loss:
\begin{align}
    \mathcal{L}_{\mathit{score~}}=\operatorname{ReLU}(\mathcal{S}'_\mathit{nm})+\operatorname{ReLU}(m-\mathcal{S}'_\mathit{mat}),
    \label{eq:score_loss}
\end{align}
where $\mathcal{S}'_\mathit{nm}$ is the non-match scores of $\mathcal{S}'$, $\mathcal{S}'_\mathit{mat}$ is the match score of $\mathcal{S}'$. Unlike the original triplet loss, this formulation provides more constraints:
\begin{itemize}
    \item Directly suppresses non-match scores ($\operatorname{ReLU}(\mathcal{S}'_\mathit{nm})$): encouraging they remain below decision thresholds.
    \item Enforces a margin on match scores ($\operatorname{ReLU}(m-\mathcal{S}'_\mathit{mat})$): guaranteeing they exceed non-matches by $m$.
\end{itemize}

By jointly optimizing score magnitudes and relative margins, the loss aligns training objectives with evaluation metrics (\eg, TAR@FAR), reducing false acceptances while maintaining discriminative power.

%% file: 4-experiments.tex
\section{Experiments} \label{sec:experiments}

To rigorously validate our method’s robustness, we intentionally leverage a diverse set of embedding models spanning multiple modalities, including face recognition model~\cite{kim2024keypoint, kim2022adaface}, gait recognition and person ReID models~\cite{ye2024biggait, liu2024distilling, gu2022clothes, wu2020adaptive, yang2023good}. This cross-modal diversity systematically avoids overfitting to any single modality’s biases, demonstrating that our framework generalizes across heterogeneous feature spaces. We stress-test our method’s ability to harmonize divergent embeddings—a critical requirement for real-world deployment, where the distribution of the test set is unpredictable.

\paragraph{Baseline Setup.} 
 We benchmark our method against traditional and contemporary fusion strategies spanning three categories: (1) \textit{Statistical Fusion}: Min/Max score fusion~\cite{jain2005score}, Z-score normalization and min-max normalization~\cite{snelick2003multimodal}; (2) \textit{Representation Harmonization}: Rank-based histogram equalization (RHE)~\cite{he2010performance}; and (3) \textit{Model-driven learnable score-fusion}: Farsight~\cite{liu2024farsight}, SVM-based (Support Vector Machine) score fusion (BSSF)~\cite{teng2022optimized},  Weighted-sum with learnable coefficients~\cite{park2021multi} and AsymA-O1’s asymmetric aggregation~\cite{herbadji2020combining}. We also compare with SapiensID~\cite{kim2025sapiensid}, a SoTA multimodal model for human recognition. This comprehensive comparison validates our method’s superiority in balancing discriminative feature preservation. 

\paragraph{Evaluation Metrics.} We adopt standard person ReID metrics like Cumulative Matching Curve (CMC) at rank-1 and mean Average Precision (mAP)~\cite{gu2022clothes, liu2024distilling}. To holistically assess whole-body biometric systems, we extend evaluation to verification (TAR@FAR: True Acceptance Rate at a False Acceptance Rate) and open-set search (FNIR@FPIR: False Non-Identity Rate at a specified False Positive Identification Rate).
\begin{itemize}
    \item TAR@FAR reflects real-world security needs: measuring reliable genuine acceptance rates while rejecting impostors within controlled error tolerance.
    \item FNIR@FPIR handles open-set scenarios (common in surveillance), rejecting unseen identities robustly without compromising known match detection.
\end{itemize}

Together, these metrics ensure that the proposed methods achieve a balanced trade-off among accuracy (CMC/mAP), security (TAR@FAR), and generalizability (FNIR@FPIR), reflecting real-world deployment requirements through a comprehensive and practical performance evaluation.

\begin{table}[t!]
\tabcolsep=0.1cm
\centering
\resizebox{\linewidth}{!}{
\begin{tabular}{lcccc}
\toprule
Dataset & Type  & \#Subjects (Train/Test/Non-mated) & \#Query & \#Gallery \\
\midrule
CCVID  & Video & 75 / 151 / 31 & 834  & 1074  \\
MEVID  & Video & 104 / 54 / 11 & 316 & 1438  \\
LTCC   & Image & 77 / 75 / 15 & 493 & 7050  \\
BRIAR  & Video & 775 / 1103 / (566, 522) & 10371  & 12264 \\
\bottomrule
\end{tabular}
}
\caption{Statistics of the evaluation set of human recognition benchmarks. BRIAR has two gallery protocols (\ie, 2 non-mated lists) for open-set search. The number of query and gallery indicate the number of images/sequences for image/video datasets.}
\label{tab:stat_datasets}
\end{table}

\paragraph{Datasets.} We evaluate our method on diverse datasets spanning static images, video sequences, multi-view captures, and cross-modal biometric data (shown in Tab.~\ref{tab:stat_datasets}) to rigorously assess generalization across varying resolutions, viewpoints, and temporal dynamics. This multi-faceted benchmarking ensures robustness to real-world challenges such as occlusion, motion blur, and sensor heterogeneity, validating practical applicability in unconstrained environments. More details are provided in the Supplementary.

\noindent \textbf{Evaluation Protocol.} For CCVID, MEVID, and LTCC, we evaluate under general conditions, as the focus of score-fusion is not only on the Clothes-Changing (CC) scenario. For BRIAR, we follow Farsight~\cite{liu2024distilling} and conduct two test settings: Face-Included Treatment, where facial images are clearly visible, and Face-Restricted Treatment, where facial images are in side view or captured from long distances.

\subsection{Implementation Details}

In our experiments, we set $N$ as either 2 or 3, incorporating multiple modalities as inputs for a comprehensive evaluation. We adopt the methodology of CAFace~\cite{kim2022cluster} to precompute gallery features for all training subjects across modalities. Specifically, pre-trained biometric backbones process all video sequences or images in the training dataset before training and use average pooling to generate modality-specific center features as gallery features. For open-set evaluation, we follow~\cite{su2025open} to construct 10 random subsets of gallery subjects which contain around $20\%$ of the subjects in the test set as the non-mated lists (numbers of non-mated subjects in Tab.~\ref{tab:stat_datasets}), and report the median and standard deviation values. During training, we randomly sample $L=8$ frames from each tracklet video and aggregate their features, either through averaging or using specific aggregation methods from the models, to produce query-level features. We set the number of experts to $Z=2$, with $p_1=w_n$, and $p_2=1-p_1$. $\delta$ is set to 3 for CCVID, MEVID, and LTCC, and 20 for BRIAR. ${\varepsilon_1, \varepsilon_2, \dots, \varepsilon_z}$ represents 3-layer MLPs. The parameter $m$ in Eq.~\ref{eq:score_loss} is set to 3. We use Adam optimizer with a learning rate of $5e^{-5}$ and a weight decay of $1e^{-2}$. We apply a Cosine annealing warm-up strategy to adjust the learning rate. For learnable baseline methods, we train them on the same training set. More details are provided in the Supplementary.

\begin{table}[t!]
\centering
\begin{subtable}{\linewidth}
\tabcolsep=0.1cm
    \centering
    \resizebox{\linewidth}{!}{
        \begin{tabular}{cccccc}
            \toprule
            \textit{Method} &\textit{Comb.}  & Rank1$\uparrow$ & mAP$\uparrow$ & TAR$\uparrow$ & FNIR$\downarrow$  \\ \midrule
            \textit{AdaFace}$^*$~\cite{kim2022adaface} & \textcolor{Peach}{$\blacklozenge$} & $94.0$ & $87.9$ & $75.7$ & $13.0\pm3.5$ \\
            \textit{CAL}~\cite{gu2022clothes}  & \textcolor{Cyan}{$\spadesuit$} & $81.4$ & $74.7$ & $66.3$ & $52.8\pm13.3$ \\ 
            \textit{BigGait}$^*$~\cite{ye2024biggait} & \textcolor{JungleGreen}{$\clubsuit$} & $76.7$ & $61.0$ & $49.7$ & $71.1\pm6.1$ \\ 
            \textit{SapiensID}~\cite{kim2025sapiensid} & \textcolor{Purple}{\ding{108}} & $92.6$ & $77.8$ & - & - \\           
            \midrule
            \textit{GEFF$^\dagger$}~\cite{arkushin2024geff} & \multirow{2}{*}{\textcolor{Peach}{$\blacklozenge$} \textcolor{Cyan}{$\spadesuit$}} & $89.4$ & $87.5$ & $84.0$ & $13.3\pm1.3$ \\  
            \textit{\textbf{Ours}} & & $\mathbf{93.3}$ & $\mathbf{89.5}$ & $\mathbf{86.9}$ & $\mathbf{11.4\pm1.5}$ \\ 
            \midrule
            \textit{Min-Fusion}~\cite{jain2005score} & \multirow{11}{*}{\textcolor{Peach}{$\blacklozenge$} \textcolor{Cyan}{$\spadesuit$} \textcolor{JungleGreen}{$\clubsuit$}} & $87.1$ & $79.2$ & $62.4$ & $48.5 \pm 8.7$ \\
            \textit{Max-Fusion}~\cite{jain2005score} & & $89.9$ & $89.3$ & $73.4$ & $23.0\pm10.1$ \\
            \textit{Z-score}~\cite{snelick2003multimodal} & & $92.2$ & $90.6$ & $73.9$ & $15.1\pm1.5$ \\
            \textit{Min-max}~\cite{snelick2003multimodal} & & $91.8$ & $90.9$ & $73.9$ & $15.4\pm2.5$ \\
            \textit{RHE}~\cite{he2010performance} & & $91.7$ & $90.2$ & $73.1$ & $16.6\pm2.5$ \\
            \textit{Weigthed-sum}~\cite{park2021multi} & & $91.7$ & $90.6$ & $73.6$ & $15.4\pm1.8$  \\
            \textit{Asym-AO1}~\cite{herbadji2020combining} & & $92.3$ & $90.0$ & $74.0$ & $15.9\pm1.7$ \\
            \textit{BSSF}~\cite{teng2022optimized} & & $91.8$ & $91.1$ & $73.9$ & $14.1\pm1.3$ \\             
            \textit{Farsight}~\cite{liu2024farsight} & & $92.0$ & $\underline{91.2}$ & $73.9$ & $13.9\pm1.1$ \\
            \rowcolor{gray!25}\textit{\textbf{Ours (AdaFace-QE)}} & & \underline{$92.6$} & $\mathbf{91.6}$ & \underline{$75.0$} & \underline{$13.3\pm1.2$} \\
            \rowcolor{gray!25}\textit{\textbf{Ours (CAL-QE)}} & & $\mathbf{94.1}$ & $90.8$ & $\mathbf{76.2}$ & $\mathbf{12.3\pm1.4}$ \\
            \bottomrule
        \end{tabular}
    }
    \caption{\bfseries Performance on CCVID Dataset.}
    \label{tab:ccvid_performance}
\end{subtable}%
\hfill
\begin{subtable}{\linewidth}
\tabcolsep=0.1cm
    \centering
    \resizebox{\linewidth}{!}{
        \begin{tabular}{ccccccc}
            \toprule
            \textit{Method} &\textit{Comb.}  & Rank1$\uparrow$ & mAP$\uparrow$ & TAR$\uparrow$ & FNIR$\downarrow$  \\ \midrule
            \textit{AdaFace}$^*$~\cite{kim2022adaface} & \textcolor{Peach}{$\blacklozenge$} & $25.0$ & $8.1$ & $5.4$ & $98.8\pm1.2$ \\
            \textit{CAL}~\cite{gu2022clothes} & \textcolor{Cyan}{$\spadesuit$} & $52.5$ & $27.1$ & $34.7$ & $67.8\pm7.3$ \\ 
            \textit{AGRL}~\cite{wu2020adaptive} & \textcolor{Cyan}{$\blacksquare$} & $51.9$ & $25.5$ & $30.7$ & $69.4\pm8.9$ \\ 
            \midrule
            \textit{GEFF$^\dagger$}~\cite{arkushin2024geff} & \multirow{2}{*}{\textcolor{Peach}{$\blacklozenge$} \textcolor{Cyan}{$\spadesuit$}} & $32.9$ & $18.8$ & $19.9$ & $78.7\pm8.1$ \\  
            \textit{\textbf{Ours}} & & $\mathbf{33.5}$ & $\mathbf{19.9}$ & $\mathbf{26.2}$ & $\mathbf{72.5\pm10.3}$ \\ 
            \midrule
            \textit{Min-Fusion}~\cite{jain2005score} & \multirow{11}{*}{\textcolor{Peach}{$\blacklozenge$} \textcolor{Cyan}{$\spadesuit$} \textcolor{Cyan}{$\blacksquare$}} & $46.8$ & $21.2$ & $28.0$ & $70.4\pm8.0$ \\
            \textit{Max-Fusion}~\cite{jain2005score}  & & $33.2$ & $14.9$ & $8.3$ & $97.4\pm1.6$ \\
            \textit{Z-score}~\cite{snelick2003multimodal} & & $54.1$ & $27.4$ & $30.7$ & $66.5\pm7.0$ \\
            \textit{Min-max}~\cite{snelick2003multimodal}  & & $52.8$ & $24.7$ & $25.0$ & $71.3\pm6.1$ \\
            \textit{RHE}~\cite{he2010performance} & & $52.8$ & $24.8$ & $25.3$ & $71.2\pm6.2$ \\
            \textit{Weigthed-sum}~\cite{park2021multi} & & $54.1$ & $27.3$ & $30.3$ & $66.3\pm7.0$ \\            
            \textit{Asym-AO1}~\cite{herbadji2020combining} & & $52.5$ & $22.9$ & $23.6$ & $71.7\pm5.8$ \\
            \textit{BSSF}~\cite{teng2022optimized} & & $53.5$ & $27.4$ & $30.5$ & $65.9\pm7.2$ \\          
            \textit{Farsight}~\cite{liu2024distilling}  & & $53.8$ & $25.4$ & $26.6$ & $69.8\pm6.4$ \\
            \rowcolor{gray!25}\textit{\textbf{Ours (AdaFace-QE)}} & & $\mathbf{55.7}$ & $\mathbf{28.2}$ & $\mathbf{32.9}$ & \underline{$64.6\pm8.2$} \\
            \rowcolor{gray!25}\textit{\textbf{Ours (CAL-QE)}} & & \underline{$55.4$} & \underline{$27.9$} & \underline{$32.5$} & $\mathbf{64.3\pm8.7}$ \\   
            \bottomrule
        \end{tabular}
    }
    \caption{\bfseries Performance on MEVID Dataset.}
    \label{tab:mevid_performance}
\end{subtable}
\caption{Our performance on CCVID and MEVID datasets in the general setting. [Keys: \textbf{Best} and \underline{second best} performance; \textit{Comb.}: model combination; $^*$: zero-shot performance; $\dagger$: reproduced using AdaFace~\cite{kim2022adaface} as the face module; \textcolor{Peach}{$\blacklozenge$}: AdaFace for face modality; \textcolor{JungleGreen}{$\clubsuit$}; BigGait for gait modality; \textcolor{Cyan}{$\spadesuit$}: CAL of body modality; \textcolor{Cyan}{$\blacksquare$}: AGRL for body modality; \textcolor{Purple}{\ding{108}}: SapiensID for face and body modality; TAR: TAR@1\%FAR; FNIR: FNIR@1\%FPIR.]
}
\label{tab:ccvid_mevid_performance}
\vspace{-1em}
\end{table}

\subsection{Experimental Results} \label{sec:lsf_performance}

Tab.~\ref{tab:ccvid_mevid_performance},~\ref{tab:ltcc_performance}, and~\ref{tab:lsf_moe_performance_evp500blended} show the performance of our method on CCVID, MEVID, LTCC, and BRIAR compared with other score-fusion methods. For Z-score and Min-max normalization methods, we average the scores after the normalization. To ensure a fair comparison with GEFF~\cite{arkushin2024geff}, we replace the FR model in GEFF with AdaFace and apply Gallery Enrichment (GE) to our method, as GE adds selected query samples into the gallery. GEFF requires a hyperparameter $\alpha$ to combine the ReID and FR score matrices and cannot extend to three modalities.

In CCVID, the FR model performs particularly well, as most body images are front-view and contain well-captured faces. As a result, the improvement through multimodal fusion is understandably limited. In MEVID, LTCC, and BRIAR (Face-Restricted Treatment), the performance of the FR model is not comparable to that of the ReID models. This is mainly due to (1) the presence of multiple views and varying distances in captured images, which often results in low-quality images, and (2) label noise and detection errors. 
The performance of score fusion surpasses that of individual models and modalities, suggesting that each model contributes complementary information. Our method effectively harnesses additional useful information in complex scenarios, leading to an even greater performance boost in MEVID and LTCC than in CCVID.
While other score-fusion approaches do not consistently perform well across all metrics or need to manually select hyperparameters, our method achieves higher performance across the board, with notable improvements in both closed-set and open-set evaluations, especially in MEVID and BRIAR. Additionally, our approach is generalizable, adapting effectively to various modality combinations, model combinations, and similarity metrics, irrespective of whether the backbones are fine-tuned on the target dataset or not. More experimental results can be found in the Supplementary.

\begin{table}[t!]
\centering
\tabcolsep=0.1cm
\resizebox{0.9\linewidth}{!}{
    \begin{tabular}{ccccccc}
        \toprule
        \textit{Method} &\textit{Comb.}  & Rank1$\uparrow$ & mAP$\uparrow$ & TAR$\uparrow$ & FNIR$\downarrow$  \\ \midrule
        \textit{AdaFace}$^*$~\cite{kim2022adaface} & \textcolor{Peach}{$\blacklozenge$} & $18.5$ & $5.9$ & $2.4$ & $99.8\pm0.2$ \\
        \textit{CAL}~\cite{gu2022clothes} & \textcolor{Cyan}{$\spadesuit$} & $74.4$ & $40.6$ & $36.7$ & $59.7\pm7.3$ \\ 
        \textit{AIM}~\cite{yang2023good} & \textcolor{Cyan}{$\blacksquare$} & $74.8$ & $40.9$ & $37.0$ & $66.2\pm9.2$ \\ 
        \textit{SapiensID}~\cite{kim2025sapiensid} & \textcolor{Purple}{\ding{108}} & $72.0$ & $34.6$ & - & - \\      
        \midrule
        \textit{\textbf{Ours}} & \textcolor{Cyan}{$\spadesuit$} \textcolor{Cyan}{$\blacksquare$} & $\mathbf{75.3}$ & $\mathbf{42.5}$ & $\mathbf{38.1}$ & $\mathbf{58.6\pm9.6}$ \\
        \midrule
        \textit{Min-Fusion}~\cite{jain2005score} & \multirow{10}{*}{\textcolor{Peach}{$\blacklozenge$} \textcolor{Cyan}{$\spadesuit$} \textcolor{Cyan}{$\blacksquare$}} & $38.1$ & $13.5$ & $12.4$ & $81.9\pm6.0$ \\
        \textit{Max-Fusion}~\cite{jain2005score}  & & $62.5$ & $33.3$ & $16.8$ & $94.8\pm4.7$ \\
        \textit{Z-score}~\cite{snelick2003multimodal} & & $73.0$ & $37.5$ & $30.4$ & \underline{$68.7\pm9.2$} \\
        \textit{Min-max}~\cite{snelick2003multimodal}  & & $73.2$ & $38.1$ & $31.9$ & $75.1\pm9.2$ \\
        \textit{RHE}~\cite{he2010performance} & & $70.4$ & $34.2$ & $21.5$ & $78.0\pm10.0$ \\
        \textit{Weigthed-sum}~\cite{park2021multi} & & $73.2$ & $37.8$ & $31.3$ & $72.4\pm8.6$ \\ 
        \textit{Asym-AO1}~\cite{herbadji2020combining} & & $71.2$ & $32.9$ & $19.1$ & $76.3\pm8.9$ &  \\
        \textit{BSSF}~\cite{teng2022optimized} & & $\underline{73.5}$ & \underline{$39.1$} & \underline{$34.2$} & $68.9\pm8.5$ \\                 
        \textit{Farsight}~\cite{liu2024farsight}  & & $73.2$ & $37.8$ & $31.3$ & $72.4\pm8.6$ \\
        \rowcolor{gray!25}\textit{\textbf{Ours}} & & $\mathbf{73.8}$ & $\mathbf{39.6}$ & $\mathbf{35.0}$ & $\mathbf{64.3\pm8.0}$ \\        
        \bottomrule
    \end{tabular}
}
\caption{Our performance on LTCC. [Keys: \textbf{Best} and \underline{second best} performance; \textit{Comb.}: model combination; $^*$: zero-shot performance; \textcolor{Peach}{$\blacklozenge$}: AdaFace for face modality; \textcolor{Cyan}{$\spadesuit$}: CAL of body modality; \textcolor{Cyan}{$\blacksquare$}: AIM for body modality; \textcolor{Purple}{\ding{108}}: SapiensID for face and body modality; TAR: TAR@1\%FAR; FNIR: FNIR@1\%FPIR.]}
\label{tab:ltcc_performance}
\vspace{-0.5em}
\end{table}

\subsection{Analysis}

Our experiments reveal two critical insights: 
\begin{enumerate}
    \item While existing methods perform well on high-quality facial datasets, they falter under challenging in-the-wild conditions characterized by non-frontal angles and variable capture quality.
    \item Our framework demonstrates superior robustness in these complex scenarios, achieving markedly larger performance gains compared to controlled environments.
\end{enumerate} 
This divergence stems from fundamental dataset characteristics: constrained benchmarks predominantly contain optimal facial captures where conventional face recognition excels, whereas unconstrained datasets reflect real-world imperfections that degrade reliability. The limitations of prior approaches arise from their dependence on high-quality facial predictions, which introduce noise when inputs diverge from ideal conditions. Conversely, our method dynamically adapts to input quality variations, synthesizing multi-modal cues to maintain accuracy without additional hardware or data requirements. This capability underscores its practical viability in deployment scenarios where sensor fidelity and environmental conditions are unpredictable.

\paragraph{Single Model Could Be Better than Fusion.}
While fusion methods generally outperform individual models, exceptions exist (\eg, LTCC), where 3-modality fusion underperforms due to weak face modality. However, fusion with CAL and AIM shows better results, serving as a direction for further mitigating such effects in future work. More results are in the Supplementary.

\paragraph{Comparison with SoTA Human Recognition Model.}  
We benchmark against SapiensID~\cite{kim2025sapiensid} on the CCVID and LTCC datasets. While SapiensID demonstrates competitive or superior performance relative to certain score-fusion methods, our method consistently achieves optimal results. This performance advantage substantiates the critical importance of score-fusion algorithm and our proposed QME.

\subsection{Ablation Studies} \label{subsec:ablation_studies}

\begin{table}[t!]
\centering
\tabcolsep=0.1cm
\resizebox{\linewidth}{!}{
    \begin{tabular}{cccccccc}
        \toprule
            \multirow{3}{*}{\textit{Method}} & \multirow{3}{*}{\textit{Comb.}}
            &\multicolumn{3}{c}{\textit{Face Incl.~Trt.}}  & \multicolumn{3}{c}{\textit{Face Restr.~Trt.}} \vspace{0.3em} \\
        \cmidrule(l{2pt}r{2pt}){3-5} \cmidrule(l{2pt}r{2pt}){6-8}
        & & TAR$\uparrow$ & R20$\uparrow$ & FNIR$\downarrow$ & TAR$\uparrow$ & R20$\uparrow$ & FNIR$\downarrow$ \\ \midrule
        \textit{KPRPE}~\cite{kim2024keypoint} & \textcolor{Peach}{$\blacklozenge$} & $66.5$ & $80.5$ & $54.8$ & $31.5$ & $44.5$ & $81.3$  \\
        \textit{BigGait}~\cite{ye2024biggait} & \textcolor{JungleGreen}{$\clubsuit$} & $66.3$ & $93.1$ & $72.7$ & $61.0$ & $90.4$ & $76.3$  \\   
        $\textit{CLIP3DReID}$~\cite{liu2024distilling} & \textcolor{Cyan}{$\spadesuit$} & $55.8$ & $83.5$ & $80.1$ & $47.9$ & $79.3$ & $83.4$  \\ \midrule
        Min-Fusion~\cite{jain2005score} & \multirow{9}{*}{\textcolor{Peach}{$\blacklozenge$} \textcolor{JungleGreen}{$\clubsuit$} \textcolor{Cyan}{$\spadesuit$}} & $70.9$ & $86.5$ & $55.6$ & $39.1$ & $58.0$ & $77.1$ \\
        \textit{Max-Fusion}~\cite{jain2005score} & & $68.7$ & $93.0$ & $72.5$ & $61.6$ & $90.6$ & $76.1$ \\
        \textit{Z-score}~\cite{snelick2003multimodal} & & $78.5$ & $92.3$ & $43.8$ & $51.1$ & $83.9$ & $72.2$ \\
        \textit{Min-max}~\cite{snelick2003multimodal} & & $82.4$ & $\mathbf{96.0}$ & $46.9$ & $61.4$ & $\mathbf{91.5}$ & $68.5$ \\
        \textit{RHE}~\cite{he2010performance} & & $82.8$ & $95.7$ & $44.2$ & $64.9$ & $90.8$ & $67.1$ \\
        \textit{Weigthed-sum}~\cite{park2021multi} & & \underline{$84.0$} & $95.4$ & $43.2$ & $62.6$ & $90.2$ & $68.1$\\ 
        \textit{Asym-AO1}~\cite{herbadji2020combining} & & $83.4$ & $95.1$ & \underline{$42.4$} & $58.5$ & $90.0$ & \underline{$66.9$} \\ 
        % \textit{BSSF}~\cite{teng2022optimized} & & $83.0$ & $95.4$ & $43.2$ & $62.6$ & $90.2$ & $68.1$\\
        \textit{Farsight}~\cite{liu2024farsight} & & 82.4 & \underline{$95.8$} & $46.1$ & \underline{$65.7$} & \underline{$91.0$} & $68.2$ \\
        \rowcolor{gray!25}\textbf{Ours} & & $\mathbf{84.5}$ & $\mathbf{96.0}$ & $\mathbf{41.2}$ & $\mathbf{67.9}$ & 90.6 & $\mathbf{64.1}$ \\
        \bottomrule
    \end{tabular}
}
    \caption{Our performance on BRIAR Evaluation Protocol 5.0.0. [Keys: \textbf{Best} and \underline{second best} performance; \textit{Comb.}: model combination; \textit{Face Incl.~Trt.}: Face-Included Treatment; \textit{Face Restr.~Trt.}: Face-Restricted Treatment; \textcolor{Peach}{$\blacklozenge$}: AdaFace for face modality; \textcolor{JungleGreen}{$\clubsuit$}: BigGait for gait modality; \textcolor{Cyan}{$\spadesuit$}: CLIP3DReID of body modality; TAR: TAR@0.1\%FAR; R20: Rank20; FNIR: FNIR@1\%FPIR.]}
    \label{tab:lsf_moe_performance_evp500blended}
\end{table}

\begin{table}[t!]
\centering
\tabcolsep=0.1cm
\resizebox{0.8\linewidth}{!}{
    \begin{tabular}{ccccccc}
        \toprule
        \textit{$\mathcal{L}_{\mathrm{score~}}$} & QE & $Z$ & Rank1$\uparrow$ & mAP$\uparrow$ & TAR$\uparrow$ & FNIR$\downarrow$ \\ \midrule
        \ding{55} & \ding{55} & 1 & $49.4$ & $21.6$ & $23.3$ & $84.0$ \\
        \ding{51} & \ding{55} & 1 & $53.8$ & $24.5$ & $25.3$ & $70.4$ \\
        \ding{55} & \ding{55} & 2 & $54.1$ & $25.5$ & $30.8$ & $65.4$ \\
        \ding{51} & \ding{55} & 2 & $55.1$ & $27.0$ & $31.3$ & $66.5$ \\
        \rowcolor{gray!25}\ding{51} & \ding{51} & 2 & $\mathbf{55.7}$ & $\mathbf{28.2}$ & $\mathbf{32.9}$ & $\mathbf{64.6}$ \\
        \bottomrule
    \end{tabular}
}
    \caption{Ablation study results on MEVID. In the absence of the QE setting (\ie, QE \ding{55}), we average the outputs from experts. [Keys: TAR= TAR@1\%FAR; FNIR= FNIR@1\%FPIR.]}
    \label{tab:ablation_study}
\vspace{-0.5em}
\end{table}

\paragraph{Effects of $\mathcal{L}_{\mathrm{score}}$, QE, and $Z$.} Tab.~\ref{tab:ablation_study} illustrates the effects of $\mathcal{L}_{\mathrm{score}}$, QE, and the number of score-fusion experts $Z$. Compared to $\mathcal{L}_{\mathrm{tri}}$, $\mathcal{L}_{\mathrm{score}}$ yields significant performance improvements across all metrics, regardless of $z$, underscoring the importance of extra boundary for non-match scores. We further observe that increasing the number of experts $Z$ gradually improves performance, indicating that combining multiple experts enriches the model’s decision-making process by capturing diverse perspectives in complex multi-modal settings. Moreover, incorporating QE guidance further boosts performance by enabling quality-aware weighting, allowing each expert to focus on the most relevant features for a given input. This reflective weighting strategy allows the experts to learn more effectively by prioritizing high-quality information, ultimately enhancing the overall robustness and accuracy of the model.

\paragraph{Effects of Mixture of Score-fusion Experts.} Tab.~\ref{tab:ablation_experts} analyzes the effects of the mixture of score-fusion experts compared to single-expert performance. 
We conduct the ablation study on BRIAR as Face Included Treatment and Face Restricted Treatment settings are closely related to face quality weights. $\varepsilon_1$ achieves better results in TAR@0.1\%FAR for Face Included Treatment and in FNIR@1\%FPIR across all settings, while $\varepsilon_2$ performs better in TAR@0.1\%FAR for Face Restricted Treatment. This is because the FR model excels in identifying true positive pairs, resulting in lower FNIR@1\%FPIR. Guided by $p_1$, $\varepsilon_1$ learns to prioritize the FR model, while $\varepsilon_2$ focuses on ReID and GR models. Fusing both experts' scores improves overall performance, demonstrating that using multiple experts enhances final performance and allows each expert to capture distinct information.

\paragraph{Effects of QE for Other Modalities.} We validate the proposed QE by evaluating the performance of QME using the QE trained from CAL as input to $\mathcal{N}_r$ in Tab.~\ref{tab:ccvid_mevid_performance} (denoted as \textit{CAL-QE}). When using QE from CAL, the performance is comparable to that of QE from AdaFace, with both significantly outperforming baseline methods. These results demonstrate the flexibility and robustness of QME.

\begin{table}[t!]
\centering
\tabcolsep=0.1cm
\resizebox{0.9\linewidth}{!}{
    \begin{tabular}{ccccccc}
        \toprule
        \multirow{2}{*}{Expert} &\multicolumn{3}{c}{\textit{Face Incl.~Trt.}}  & \multicolumn{3}{c}{\textit{Face Restr.~Trt.}} \vspace{0.3em} \\
        \cmidrule(l{2pt}r{2pt}){2-4} \cmidrule(l{2pt}r{2pt}){5-7}
        & TAR$\uparrow$ & R20$\uparrow$ & FNIR$\downarrow$ & TAR$\downarrow$ & R20$\uparrow$ & FNIR$\downarrow$ \\ \midrule
        $\varepsilon_1$ & \underline{83.6} & \underline{95.5} & \underline{41.7} & 62.0 & $\mathbf{90.6}$ & \underline{66.7} \\
        $\varepsilon_2$ & 81.8 & \underline{95.5} & 46.6 & \underline{65.0} & $\mathbf{90.6}$ & 68.4  \\
        \rowcolor{gray!25}Ours ($\varepsilon_1+\varepsilon_2$) & $\mathbf{84.5}$ & $\mathbf{95.7}$ & $\mathbf{41.2}$ & $\mathbf{67.9}$ & $\mathbf{90.6}$ & $\mathbf{64.1}$ \\
        \bottomrule
    \end{tabular}
}
    \caption{Effects of the mixture of score-fusion experts on BRIAR. $\varepsilon_1$ has a better performance in \textit{Face Incl.~Trt.}, while $\varepsilon_2$ experts in \textit{Face Restr.~Trt.}. [Keys: \textit{Face Incl.~Trt.}= Face Included Treatment; \textit{Face Restr.~Trt.}= Face Restricted Treatment; TAR=TAR@0.1\%FAR; R20=Rank20; FNIR=FNIR@1\%FPIR.]}
    \label{tab:ablation_experts}
\end{table}
\vspace{-0.5em}
\subsection{Visualization}

\begin{figure}[t!]
    \centering
    \includegraphics[width=0.95\linewidth]{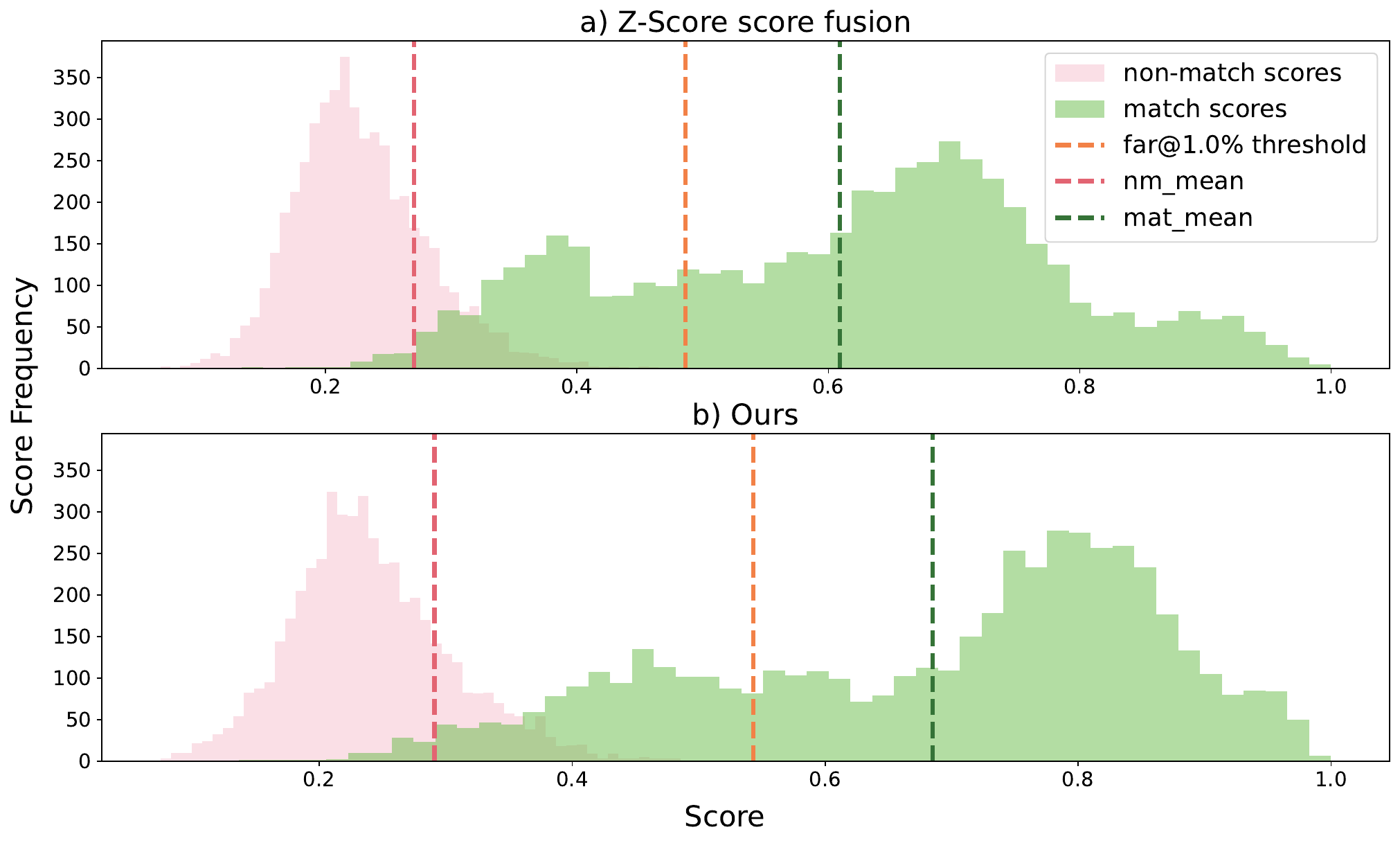}
    \caption{Score distributions of the CCVID test set. [Keys: nm\_mean = mean value of non-match scores; mat\_mean = mean value of match scores.]}
    \label{fig:scores_distribution}
\vspace{-0.5em}
\end{figure}

\paragraph{Score Distribution.} Fig.~\ref{fig:scores_distribution} visualizes the distribution of non-match scores, match scores, and the threshold FAR@1\% for both Z-score and our method on CCVID. To ensure a balanced comparison between the two distributions, we randomly sample an equal number of non-match and match scores.  Compared to the Z-score score-fusion, our approach boosts match scores while keeping non-match scores within the same range. This adjustment validates the effects of score triplet loss to improve the model’s ability to distinguish between matches and non-matches.

\begin{figure}[t!]
    \centering
    \includegraphics[width=0.95\linewidth]{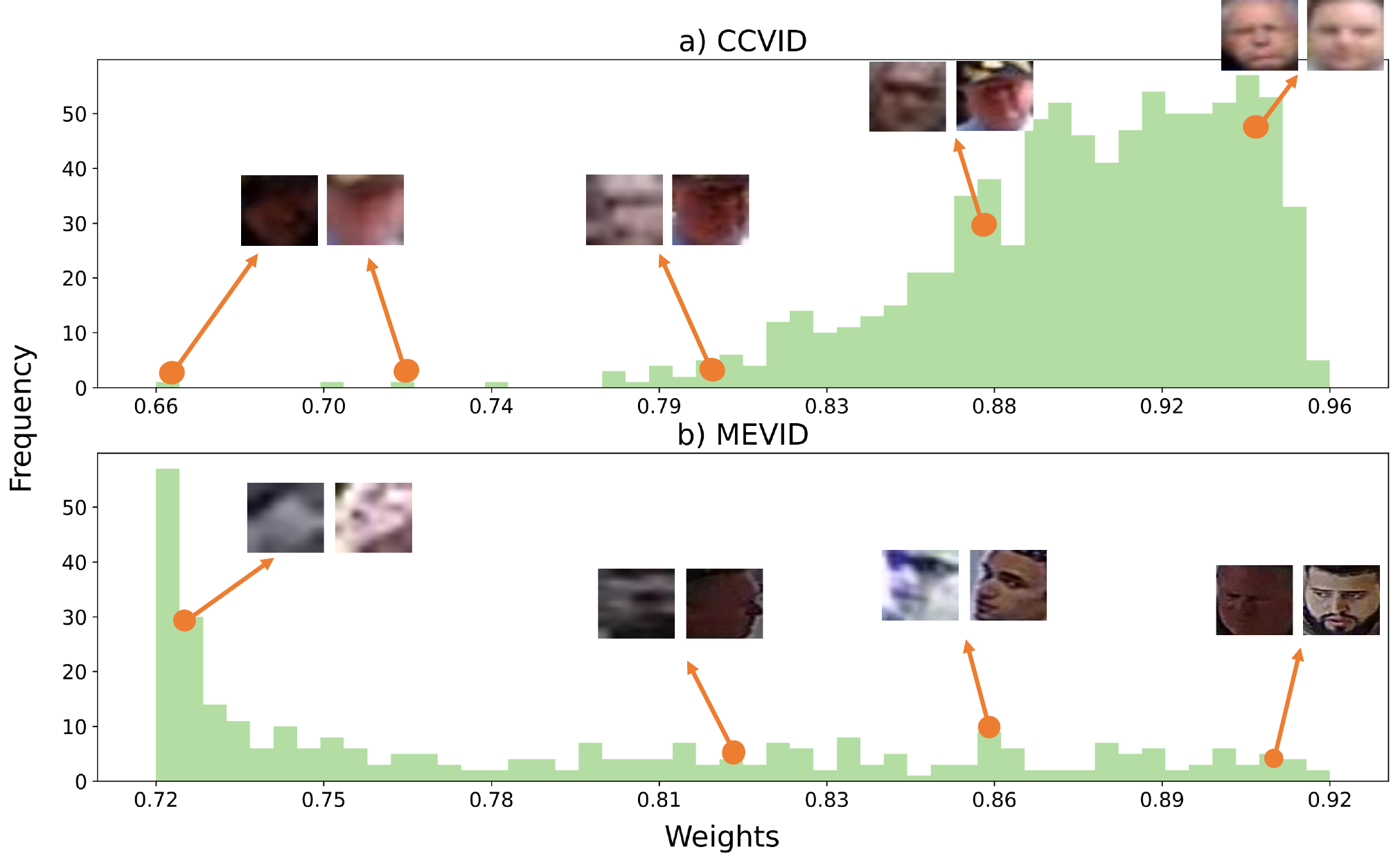}
    \caption{The distribution of AdaFace quality weights for the CCVID and MEVID datasets, illustrated with examples showcasing a range of quality weights.}
    \label{fig:qe_visualization}
\vspace{-0.5em}
\end{figure}

\paragraph{Quality Weights.} Fig.~\ref{fig:qe_visualization} visualizes the distribution of predicted quality weights for facial images in the CCVID and MEVID test sets. Note that these weights represent video-level quality weights, obtained by averaging the quality weights of each frame in the video sequence. CCVID has a higher proportion of high-quality weights, as most images are captured from a front view. In contrast, MEVID shows more variability in quality weights due to detection noise and varying clarity. The visualization indicates that our method effectively estimates image quality. The use of ranking-based pseudo-labels encourages the model to focus on relative quality, making it more robust to outliers. This guides the score-fusion experts to prioritize the most reliable modality based on quality. Visualization of CAL quality weight can be found in the Supplementary.

